\documentclass[journal]{IEEEtran} 
\usepackage{authblk}

\usepackage{graphics} 
\usepackage{graphicx}
\usepackage{array}
\usepackage{amsmath} 
\usepackage{amssymb}  
\usepackage[export]{adjustbox}
\usepackage[caption=false]{subfig}
\usepackage{algorithmic}
\usepackage[ruled]{algorithm}
\usepackage{color}
\usepackage{hyperref}
\usepackage{wrapfig}
\usepackage{mathptmx}
\usepackage{mathrsfs}
\usepackage{placeins}
\usepackage[shortcuts]{extdash}
\usepackage{siunitx}
\usepackage{numprint}
\usepackage{xspace}
\DeclareMathSymbol{\shortminus}{\mathbin}{AMSa}{"39}

\def\bfm#1{{\bf #1}}

\definecolor{fgred}{rgb}{0.8,0,0}     
\definecolor{fgorange}{rgb}{0.93,0.53,0.18}     
\definecolor{fgpurple}{rgb}{0.55,0.1,0.6}     
\definecolor{fggreen}{rgb}{0,0.5,0}     
\definecolor{fgblue}{rgb}{0,0,0.7}     
\definecolor{fgclay}{rgb}{0.51,0.25,0.04}     

\newcommand{\BB}{\textit{Blinky Block}\xspace}
\newcommand{\BBs}{\textit{Blinky Blocks}\xspace}
\newcommand{\VS}{\textit{VisibleSim}\xspace}
\usepackage{url}

\usepackage{enumitem}

\begin{document}
\title{Distributed prediction of unsafe reconfiguration scenarios of modular robotic Programmable Matter}
\author[1]{Beno\^{i}t Piranda}
\author[2]{Pawe\l{} Chodkiewicz}
\author[3]{Pawe\l{} Ho\l{}obut}
\author[4]{St\'ephane P.A. Bordas}
\author[1]{Julien Bourgeois}
\author[3,4]{Jakub Lengiewicz}
\affil[1]{University of Bourgogne Franche-Comt\'e, FEMTO-ST Institute, CNRS, France} 
\affil[2]{Faculty of Automotive and Construction Machinery Engineering\\
Warsaw University of Technology, Poland}
\affil[3]{Institute of Fundamental Technological Research\\
Polish Academy of Sciences, Poland}
\affil[4]{Department of Engineering\\ 
Faculty of Science, Technology and Medicine\\
University of Luxembourg}

\maketitle

\begin{abstract}
We present a distributed framework for predicting whether a planned reconfiguration step of a modular robot will mechanically overload the structure, causing it to break or lose stability under its own weight. The algorithm is executed by the modular robot itself and based on a distributed iterative solution of mechanical equilibrium equations derived from a simplified model of the robot. The model treats inter-modular connections as beams and assumes no-sliding contact between the modules and the ground. We also provide a procedure for simplified instability detection. The algorithm is verified in the Programmable Matter simulator \VS, and in real-life experiments on the modular robotic system \BBs.

\vspace{0.2em}\noindent\emph{Keywords:} Self-reconfiguration, Modular robots, Distributed algorithms, Mechanical constraints, Programmable Matter.

\end{abstract}

\section{Introduction}

Materials able to autonomously assume any shape are the dream of engineers. Currently, the most advanced artificial systems possessing elements of this functionality are modular self-reconfigurable robots---machines composed of robotic units (\emph{modules}) which can bond together, move over one another and communicate, as well as store and process information \cite{YimShen2007}. They may be compared to swarms of fire ants forming bio-mechanical structures from their own bodies \cite{MlotTovey2011}.
Cooperation of millions of tiny, densely-packed modules is expected to produce the desired emergent shape change of the entire ensemble. A system of this kind would be a realization of the futuristic concept of \emph{Programmable Matter}~\cite{GoldsteinCampbellMowry06}. 

The operation of densely-packed self-reconfigurable robots is based on the movement of modules from one part of the robot to another. This poses not only the challenging hardware problem of designing miniaturized modules able to operate in large 3D ensembles, but also the software problem of controlling this motion. If the robot is autonomous, its operation must be collectively planned and controlled by the modules, taking into account \emph{geometric} and \emph{mechanical} constraints on reconfiguration at each stage of motion. Although offline approaches could be used to check these constraints efficiently for a number of selected configurations, the less efficient online approaches must be used in general to handle any shape or any possible interaction with obstacles. 

Geometric constraints result from the fact that modules need sufficient space to make a planned move but also must constantly remain in physical contact with other modules \cite{Nguyen00}. Mechanical constraints, in turn, result from the requirements of integrity and stability of the entire robot---the structure cannot break at inter-modular junctions or lose balance during reconfiguration. The mechanical constraints can be neglected in some special cases, like reconfiguration under no external loading (weightlessness) or 2D reconfiguration on flat ground with perpendicular gravity as the only loading. Otherwise, they usually need to be considered.

Currently, almost all algorithms for planning and controlling self-re\-con\-fi\-gu\-ra\-tion of densely-packed systems take only geometric constraints into account, like in \cite{FitchButler2008, Stoy06, Piranda16, NazPirandaBourgeoisGoldstein16, LengHolob2019}. An up-to-date survey of self-reconfiguration algorithms for modular robots can be found in \cite{THALAMY2019}. 
By contrast, works on the mechanical behavior of densely-packed modular structures are few, present mostly centralized procedures, and rarely discuss reconfiguration. Examples can be found in \cite{White11}, focused on optimizing the compliance of modular tools, and in \cite{Hiller14}, aimed at predicting the mechanical behavior of structures produced by additive manufacturing. As a separate research direction, special types of modular structures were investigated in \cite{CampbellPillai08}, \cite{Holobut15} and \cite{Lengiewicz17} to check the possibility of using modular robots as collective actuators. Autonomous reconfiguration planning that takes into account both geometric and mechanical constraints remains a challenging open problem, some aspects of which are investigated in this work.

In the present paper, we develop the approach introduced in~\cite{Holobut17} much further into a more realistic framework.
Arbitrary 3D structures are investigated, for which a linear-elastic FE model is again adopted, with the addition of unilateral contact conditions  which represent interaction with the surroundings. Two failure modes are considered: overloading of inter-modular connections and loss of balance, both checked in a distributed manner. Two methods of checking the loss of balance are proposed: (1) a simplified one, valid for structures standing on a flat surface, and (2) a model-based one, which is more general but requires solution of the mechanical balance equations with contact conditions. Verification is performed in a dedicated simulator \VS \cite{dagstuhlVS}, as well as experimentally on the real robotic modules \BBs~\cite{GoldsteinCHI2011}. The computational cost and several possible extensions of the applied weighted Jacobi iterative solver are also discussed.

\section{Distributed prediction of the mechanical state of a robot%
}
\label{sec: distributed prediction}

\subsection{Basic characteristics of a modular robot}

For the purposes of algorithmic mechanical analysis, a modular robot will be represented only by its connection topology and inter-modular connection strengths. We will focus on structures built of cubic \BBs~\cite{GoldsteinCHI2011}, which are arranged on the Cartesian grid. However, in principle, the proposed approach can handle other connection topologies.  

From the information-theoretic point of view, a modular robot has a distributed and asynchronous computing architecture, with mesh connection topology corresponding to the communication network of modules~\cite{Tanenbaum06}. In \BBs, information can only be directly transferred between adjacent modules,  
so the communication network has the same connection topology as the modular robot itself. Such neighbor-to-neighbor-only communication imposes strict limits on the information-passing abilities of the system and requires special algorithms for effective prediction of mechanical failures.

\subsection{Problem to be solved}

Reconfiguration of a modular robot can be broken down into simple steps. A single step consists of (a) releasing some inter-modular connections, (b) shifting modules which have been freed into nearby positions and, finally, (c) creating new connections at the new locations. As an alternative way of restructuring one may consider attaching new modules to a structure and discarding some old ones. Each reconfiguration step can potentially cause failure of the structure, which is in general irreversible. To avoid it, a prior mechanical analysis can be performed to ascertain that the planned new configuration is mechanically safe. 

We restrict further presentation only to the case of attaching additional modules to an existing structure. Nevertheless, the whole idea can also be applied to a complete reconfiguration step. In such a case, the mechanical analysis would need to be performed for each of the above stages (a), (b) and (c) of the planned reconfiguration step. The step would be considered permissible if the structure was predicted to be safe after each of the three stages. Since this kind of reconfiguration cannot be easily performed on \BBs, we do not further develop this topic in the present paper.

\begin{figure}
\centering
\subfloat[]{\includegraphics[width=0.18\textwidth]{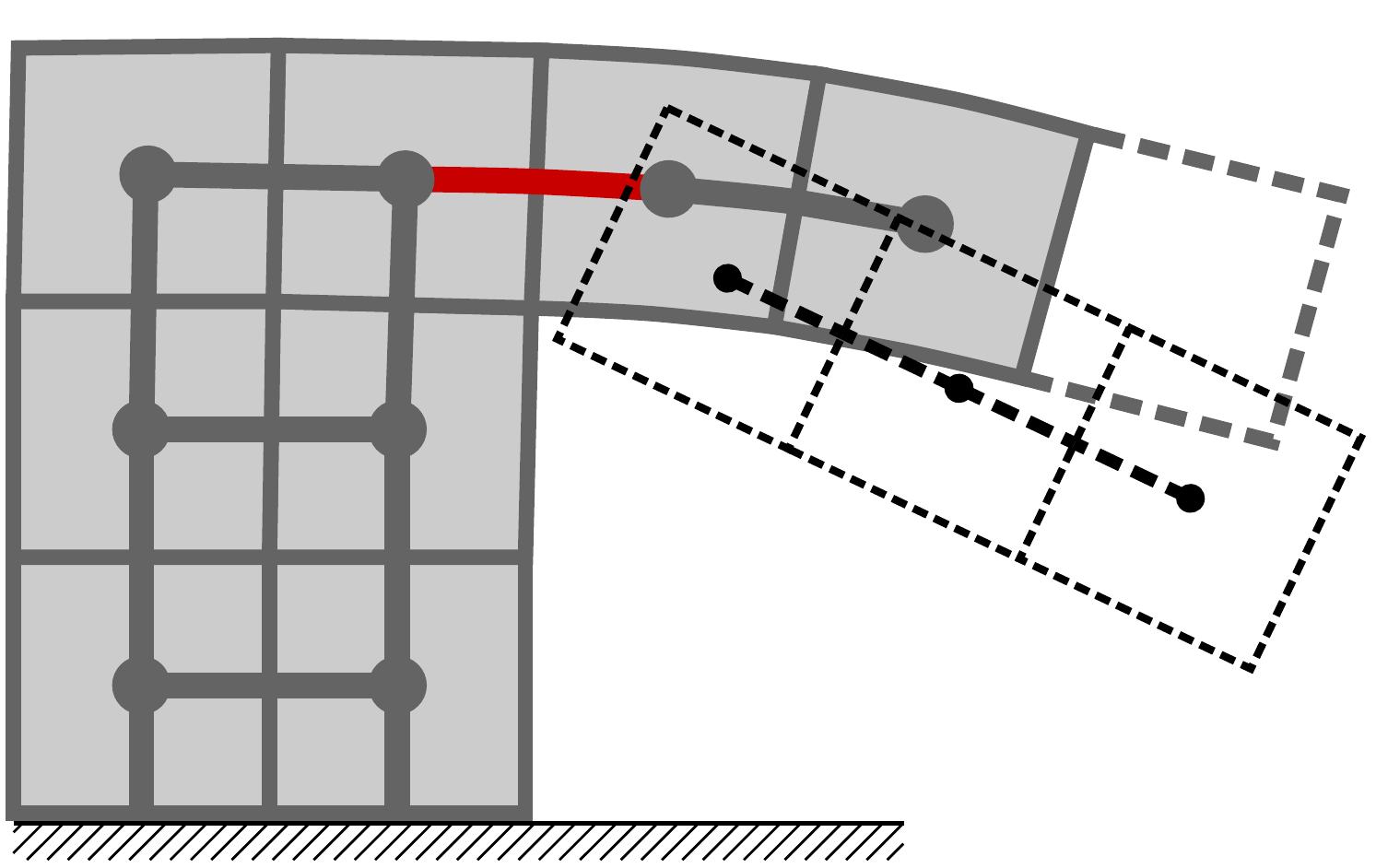}} \hfil
\subfloat[]{\includegraphics[width=0.13\textwidth]{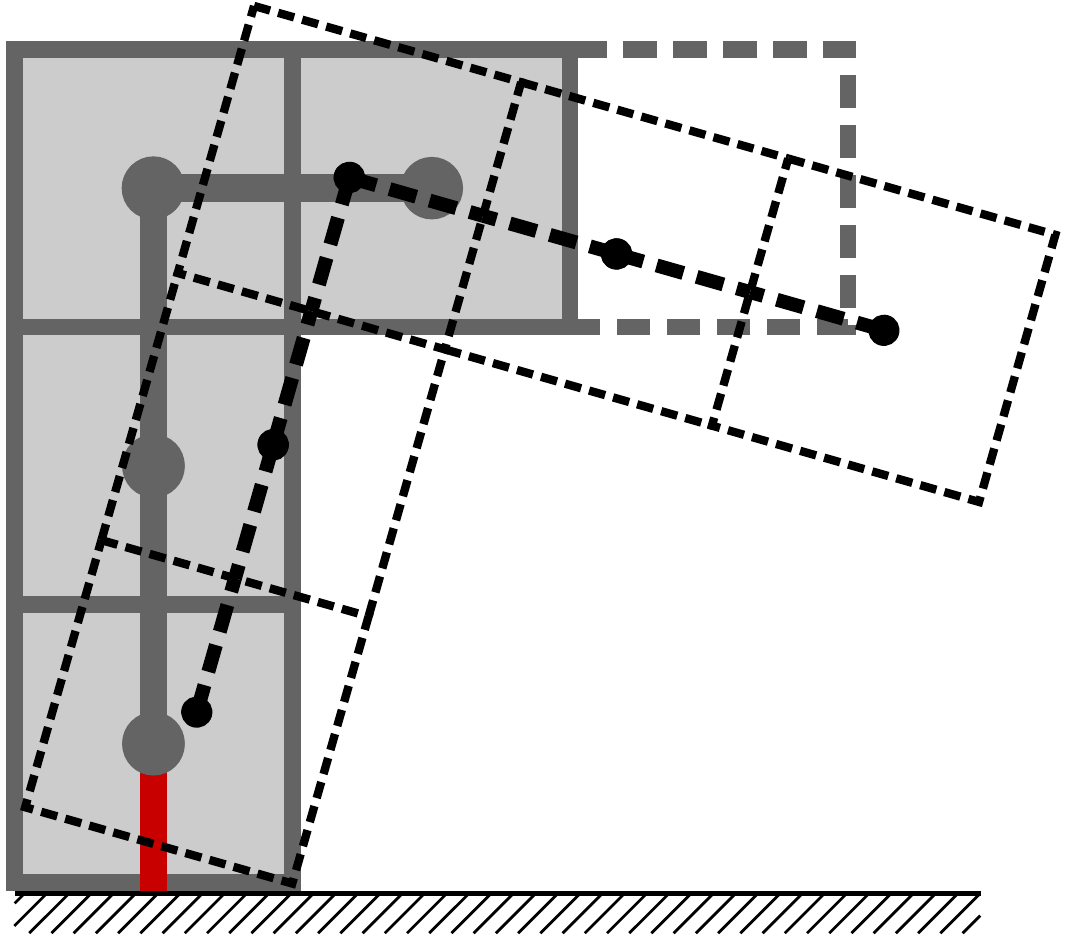}}
\caption{Two types of failure after an additional module (gray dashed line) is attached: (a) breakage of a connection; (b) loss of overall stability. The red lines designate failing connections.}
\label{fig:failures}
\end{figure}

Algorithmic prediction of two failure modes, shown in Fig.~\ref{fig:failures}, will be analyzed, expanding on the ideas proposed in~\cite{Holobut17}. The first one is when addition of new blocks increases stresses at some inter-modular junctions beyond the holding capacity of connectors, as a result of which the structure breaks, Fig.~\ref{fig:failures}a. The second one is when the structure loses balance when the new modules are added, Fig.~\ref{fig:failures}b. 

\subsection{Overview of model-based failure prediction}

We propose a distributed procedure which can predict both types of failure simultaneously. It approximately solves a special mechanical problem for the robot with new modules attached and can be customized to handle different module designs. The procedure has several distinctive features, described in detail in the succeeding sections:
\begin{itemize}
    \item The modular robot is represented by a Finite Element (FE) model; see Sec.~\ref{sec: Standard 3D frame model}.
    \item Two types of connections are assumed: the inter-modular connection, modeled as a linear-elastic beam, and the connection between a module and an external support (e.g., the ground), modeled as a linear-elastic beam with unilateral no-sliding contact conditions at the support end. 
    \item The mechanical state of a planned configuration with new modules attached (see Sec.~\ref{sec: virtual modules}) is obtained by solving a non-linear problem (non-linearities result from the contact conditions; see Sec.~\ref{sec: contact conditions}). 
    \item The problem is solved in a distributed fashion using the weighted Jacobi iterative scheme (see Sec.~\ref{sec: weighted Jacobi}).
    \item After the iterations converge, two failure criteria are checked: for the loss of balance (see Sec.~\ref{sec: stability for general case}) and overloading of connections (see Sec.~\ref{sec: overloading condition}).
\end{itemize}

\subsection{Standard 3D frame model}
\label{sec: Standard 3D frame model}

In the adopted FE model, each module $p$ is represented by a node with 6 degrees of freedom $\bfm{u}_p$ (3 displacements, $u_x, u_y, u_z$, and 3 rotations, $\tau_x, \tau_y, \tau_z$) and each pair of connected modules is represented by a beam joining their nodes. A module in contact with the ground is represented by a special beam between the module's node and a ``ground'' node $g$, with $\bfm{u}_g=\bfm{0}$, as it is explained in Sec.~\ref{sec: contact conditions}. In Fig.~\ref{fig:failures}, the beams are presented as lines joining the centers of adjacent modules. 

In a coordinate system CS whose $z$ axis points upwards, the stiffness matrices for a beam joining module $p$ and module $q$ lying below it read $\bfm{K}_{pq}^{11}=\bfm{K}^{11}$ and $\bfm{K}_{pq}^{12}=\bfm{K}^{12}$, where:
	\begin{eqnarray}
	&	\hspace{-1.2em}\bfm{K}^{11}=\dfrac{E}{L^3}\begin{pmatrix}%
 12 I_x & 0 & 0 & 0 & -6 I_x L & 0 \\
 0 & 12 I_y & 0 & 6 I_y L & 0 & 0 \\
 0 & 0 & A L^2 & 0 & 0 & 0 \\
 0 & 6 I_y L & 0 & 4 I_y L^2 & 0 & 0 \\
 -6 I_x L & 0 & 0 & 0 & 4 I_x L^2 & 0 \\
 0 & 0 & 0 & 0 & 0 & J_{\nu} L^2
	\end{pmatrix},
	\end{eqnarray}
	\begin{eqnarray}
	&	\hspace{-1.2em}\bfm{K}^{12}=\dfrac{E}{L^3}\begin{pmatrix}%
 -12 I_x & 0 & 0 & 0 & -6 I_x L & 0 \\
 0 & -12 I_y & 0 & 6 I_y L & 0 & 0 \\
 0 & 0 & -A L^2 & 0 & 0 & 0 \\
 0 & -6 I_y L & 0 & 2 I_y L^2 & 0 & 0 \\
 6 I_x L & 0 & 0 & 0 & 2 I_x L^2 & 0 \\
 0 & 0 & 0 & 0 & 0 & -J_{\nu} L^2
	\end{pmatrix},
	\end{eqnarray}
while $E$, $L$, $A$, $I_x$, $I_y$ and $J_{\nu}$ are the elastic modulus, length, cross-sectional area, area moments of inertia in the $x$ and $y$ directions, and scaled torsion constant in the $z$ direction of the beam, respectively (see Tab.~\ref{tab: BlinkyBlock material parameters}). If the two neighbors $p$ and $q$ are aligned with the $z$ axis of another coordinate system CS', then $\bfm{K}_{pq}^{11}$ and $\bfm{K}_{pq}^{12}$ take the form:
\begin{equation}
	\bfm{K}_{pq}^{11} = \bfm{\hat{R}}_{pq} \bfm{K}^{11} \bfm{\hat{R}}_{pq}^{\text{T}} \quad , \quad
	\bfm{K}_{pq}^{12} = \bfm{\hat{R}}_{pq} \bfm{K}^{12} \bfm{\hat{R}}_{pq}^{\text{T}} \,,
\end{equation}
where 
\begin{equation}
    \bfm{\hat{R}}_{pq}=
    \begin{pmatrix}
 	\bfm{R}_{pq} & \bfm{0} \\
 	\bfm{0} & \bfm{R}_{pq}
 	\end{pmatrix},
\end{equation}
$\bfm{R}_{pq}$ is the $3\times 3$ rotation matrix from CS' to CS, $\Box^\text{T}$ denotes a transpose, and block matrix notation is used.

The ensemble is in static equilibrium if, for every module $p$, the sums of reaction forces and torques between $p$ and all its neighbors $q$ are equal to the gravitational force and null external torque acting on $p$, respectively, given by the vector $\bfm{F}_p^{\text{ext}}=[0, 0, \shortminus 9.81\cdot M, 0, 0, 0]^{\text{T}}$. The equilibrium equations are:
	\begin{eqnarray}
	\label{eqn:local set of eqs}
	\forall_p \left[\sum_q \bfm{K}_{pq}^{11}\bfm{u}_p + \bfm{K}_{pq}^{12}\bfm{u}_q\right]=\bfm{F}_p^{\text{ext}} \,,
	\end{eqnarray}
with unknown vectors $\bfm{u}_p$ and $\bfm{u}_q$. After assembling the global vectors $\bfm{u}$, $\bfm{F}^{\text{ext}}$ and stiffness matrix $\bfm{K}$, incorporating all degrees of freedom of the structure, Eq.~(\ref{eqn:local set of eqs}) takes the form $\bfm{K} \bfm{u}=\bfm{F}^{\text{ext}}$. It still needs to be extended to account for the modules planned to be added (Sec.~\ref{sec: virtual modules}) and contact conditions (Sec.~\ref{sec: contact conditions}). The respective quantities for the extended system, called the perturbed state, will be denoted by symbols with bars.

\begin{table}
    \centering
    \begin{tabular}{l|l|p{0.21\textwidth}}
        symbol & value & description \\
        \hline
         $E$ & \SI{100}{\mega \pascal} & elastic modulus\\[0.2em]
         $L,\quad A=L^2$ & \SI{40}{\milli \meter},  $40\times{}40~$mm$^2$& length \& cross-sectional area\\[0.2em]
         $I_x$, $I_y$ & $L^4/12~$mm$^4$ & $x$ and $y$ moments of inertia \\[0.2em]
         $J_{\nu}=\frac{J}{2(1+\nu)}$ & $2.25\cdot L^4/41.6~$mm$^4$ & scaled $z$ torsion constant ($J$: $z$ torsion constant, $\nu$: Poisson's ratio) \\[1.4em]
         $M$ & \SI{0.06106}{\kilogram} & mass of a block \\[0.2em]
         $F^{\text{max}}_{\text{V}}$ & \SI{11.98}{\newton} & strength of a vertical connection \\[0.2em]
         $F^{\text{max}}_{\text{L}}$ & \SI{14.97}{\newton} & strength of a lateral connection
    \end{tabular}\vspace{0.5em}
    \caption{Blinky Blocks' geometric and material parameters}
    \label{tab: BlinkyBlock material parameters}
\end{table}

\subsection{Adding virtual modules}
\label{sec: virtual modules}

To predict the state of the system one reconfiguration step ahead, the algorithm must take into account \emph{virtual modules}---the modules which are planned to be attached. This is done in a simple way: a system analogous to Eq.~(\ref{eqn:local set of eqs}) is built assuming that virtual modules are present. During computation, virtual modules are emulated by their existing neighbors, which store and process variables and messages related to virtual modules.

\subsection{Contact with external supports}
\label{sec: contact conditions}

We use a simplified contact model of a cubic module with an external support, in which the support must be flat and co-planar with one of the module's facets. Also, we only analyze initially existing contact interfaces and assume that no new ones appear under load. We distinguish two conditions: (i) a unilateral contact-separation condition (coaxial mode) combined with no-sliding/no-twisting (shear and torsional modes), (ii) a tilting condition (bending mode). We say that a module is in contact if the axial force in the beam representing the contact is compressive. Otherwise, the module is in separation. When in contact, the module can only tilt if the bending torque in the beam exceeds a limit torque which is proportional to the compressive force (see the explanation below).

\begin{figure}
\centering
\subfloat{\includegraphics[width=0.49\textwidth]{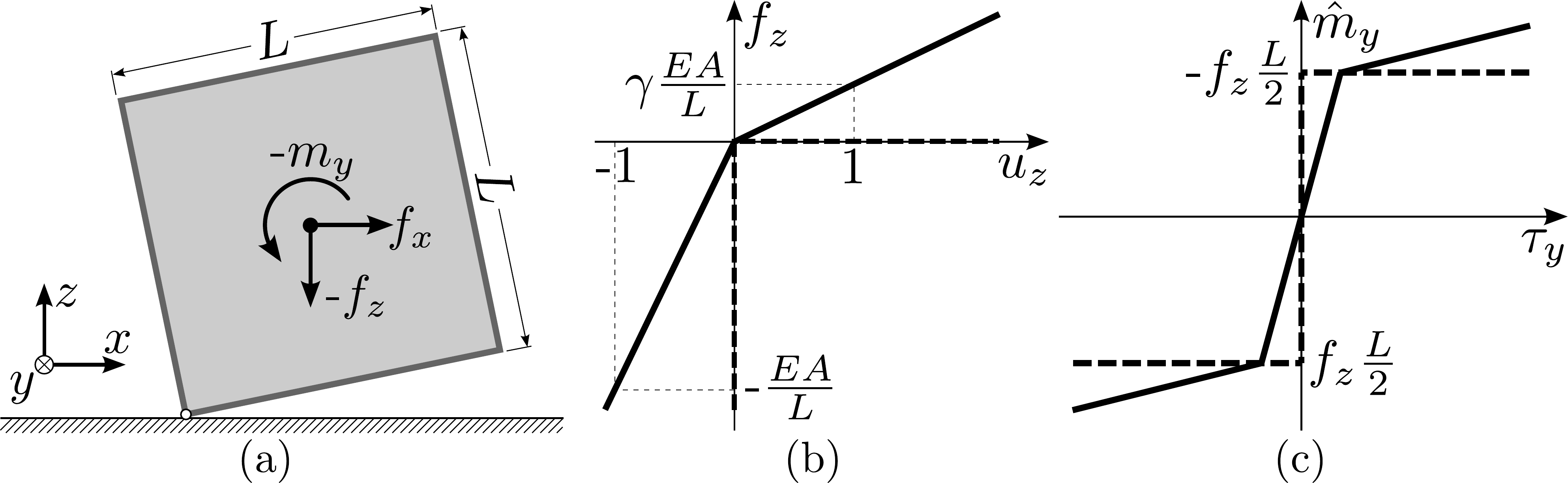}}
\caption{Regularized contact conditions: (a) a module in contact with the ground, (b) contact-separation condition for the $f_z$ component, and (c) tilting (stable/unstable) condition for the $m_y$ component. The dashed lines indicate the respective ``exact'' (non-regularized) relationships for the case of a rigid module in contact with rigid ground.}
\label{fig:regularized-contact-relationships}
\end{figure}

Without loss of generality, let us consider a module supported from below. The module is loaded with the forces and torques $\bfm{F}=[f_x, f_y, f_z, m_x, m_y, m_z]^{\text{T}}$, which correspond to the kinematic variables $\bfm{u}=[u_x, u_y, u_z, \tau_x, \tau_y, \tau_z]^{\text{T}}$. The contact condition (see Fig.~\ref{fig:regularized-contact-relationships}b) takes the form of the Signorini problem:
\begin{equation}
    \label{eq: contact-separation exact}
    f_z\leq 0\qquad \& \qquad u_z\geq 0 \qquad \& \qquad f_z u_z = 0.
\end{equation}
Additionally, we require that when the contact is active ($f_z<0$), there is no tangential slip ($u_x=0$ $\&$ $u_y=0$) or twisting ($\tau_z=0$), and the module can only tilt if at least one of the bending torques exceeds a limit torque. The tilting conditions (see Fig.~\ref{fig:regularized-contact-relationships}c) can be conveniently written in the following form:
\begin{eqnarray}
    \label{eq: stable-tilting exact x}
    & \nonumber\Phi_x \leq 0 \quad \& \quad \tau_x\cdot\mathrm{sign}(\hat{m}_x)\geq 0 \quad \& \quad \Phi_x\cdot \tau_x=0 \,,\\
    & \mathrm{where} \quad \hat{m}_x=m_x-f_y\cdot L/2 \,, \quad \Phi_x=|\hat{m}_x|+f_z\cdot L/2 \,; \\
    \label{eq: stable-tilting exact y}
    & \nonumber\Phi_y \leq 0 \quad \& \quad \tau_y\cdot\mathrm{sign}(\hat{m}_y)\geq 0 \quad \& \quad \Phi_y\cdot \tau_y=0 \,,\\
    & \mathrm{where} \quad \hat{m}_y=m_y+f_x\cdot L/2 \,, \quad \Phi_y=|\hat{m}_y|+f_z\cdot L/2 \,;
\end{eqnarray}
where $L$ is the module's size. The tilting (bending) condition expresses the fact that the torques $|\hat{m}_x|$ or $|\hat{m}_y|$ cannot exceed the torque $\shortminus f_z\cdot L/2$ produced by the compressive force $\shortminus f_z$ about any of the facet's edges; see also Fig.~\ref{fig:regularized-contact-relationships}a. 

A supported module is modeled as a beam with the same elastic constants as two connected modules. This gives the force-displacement relationship for the case of stable contact (when the module adheres to the support with an entire facet). 
This relationship is then used by a special predictor-corrector scheme, described below, which provides regularization to the contact conditions given by Eqs.~(\ref{eq: contact-separation exact}--\ref{eq: stable-tilting exact y}). 

In the regularized contact scheme for module $p$, a trial state is computed first, assuming a linear-beam-type connection with each support $q$ (here, the contact direction is $z$):
\begin{equation}
    \bfm{F}_{pq}^{\mathrm{tr}}=[f_x^{\mathrm{tr}}, f_y^{\mathrm{tr}}, f_z^{\mathrm{tr}}, m_x^{\mathrm{tr}}, m_y^{\mathrm{tr}}, m_z^{\mathrm{tr}}]^{\text{T}} = \bfm{K}^{11}_{pq}\bar{\bfm{u}}_p\,,
\end{equation}
where the term $\bfm{K}^{12}_{pq} \bar{\bfm{u}}_q$ is absent because the support is assumed to be immobile, $\bar{\bfm{u}}_q=\bfm{0}$. Then, a corrected 
vector $\bfm{F}_{pq}$ is determined, 
taking into account two conditions:\\
\emph{(i) the unilateral (normal) contact-separation condition combined with no-sliding and no-twisting conditions}: 
\begin{equation}
    \label{eq: normal contact regularized}
    \begin{cases}
    \begin{array}{llr}
    \bfm{F}_{pq}:=\gamma\cdot\bfm{F}_{pq}^{\mathrm{tr}} & \textrm{  for } f_z^{\mathrm{tr}}\geq 0 & \textrm{  (separation)}\\
    (f_x, f_y, f_z, m_z)=(f_x^{\mathrm{tr}}, f_y^{\mathrm{tr}}, f_z^{\mathrm{tr}}, m_z^{\mathrm{tr}}) & \textrm{  otherwise} & \textrm{  (contact)}
    \end{array}
    \end{cases}
\end{equation}
\noindent\emph{(ii) the tilting (bending) condition, used only when $f_z^{\mathrm{tr}}<0$, i.e. when the module is in contact:}
\begin{equation}
    \label{eq: tilting condition x regularized}
    m_x:=
    \begin{cases}
    \begin{array}{lll}
    m_x^{\mathrm{tr}} & \hspace{0.2em} \textrm{for } \Phi_x^{\mathrm{tr}}<0 & \text{(stable)}\\
    \gamma\cdot m_x^{\mathrm{tr}}+\bar{\gamma}\cdot\frac{L}{2}\left[f_y^{\mathrm{tr}}-\mathrm{sign}\left(\hat{m}_x^{\mathrm{tr}}\right)f_z^{\mathrm{tr}}\right] & \hspace{0.2em} \textrm{otherwise} & \text{(tilting)}
    \end{array}
    \end{cases}
\end{equation}
\begin{equation}
    \label{eq: tilting condition y regularized}
    m_y:=
    \begin{cases}
    \begin{array}{lll}
    m_y^{\mathrm{tr}} & \hspace{0.2em} \textrm{for } \Phi_y^{\mathrm{tr}}<0 & \text{(stable)}\\
    \gamma\cdot m_y^{\mathrm{tr}}-\bar{\gamma}\cdot\frac{L}{2}\left[f_x^{\mathrm{tr}}+\mathrm{sign}\left(\hat{m}_y^{\mathrm{tr}}\right)f_z^{\mathrm{tr}}\right] & \hspace{0.2em} \textrm{otherwise} & \text{(tilting)}
    \end{array}
    \end{cases}
\end{equation}
where $\hat{m}_x^{\mathrm{tr}}$, $\hat{m}_y^{\mathrm{tr}}$, $\Phi_x^{\mathrm{tr}}$ and $\Phi_y^{\mathrm{tr}}$ are computed as in Eqs.~(\ref{eq: stable-tilting exact x}) and (\ref{eq: stable-tilting exact y}), but using the components of $\bfm{F}_{pq}^{\mathrm{tr}}$; $\gamma=10^{-4}$ and $\bar{\gamma}=1-\gamma$.

The corrected contact tangent matrix $\bar{\bfm{K}}{}^{11}_{pq}$ is obtained as a derivative of $\bfm{F}_{pq}$ with respect to $\bar{\bfm{u}}_p$. Again, the matrix $\bar{\bfm{K}}{}^{12}_{pq}$ is disregarded because supports are assumed to be immobile.

\emph{Remark:} Ill-posedness of the problem is avoided by introducing a very weak spring, characterized by $\gamma$, that prevents free rigid-body motion of the structure. The drawback of this approach is poor conditioning of the resulting system of equations when the robot is unstable, which can deteriorate the convergence rate of the iterative solver; see Sec.~\ref{sec:convergence_properties}. However, as it is shown in Sec.~\ref{sec: stability for general case}, the knowledge of which supports are active suffices for assessing stability and those are usually identified long before convergence is achieved.

\subsection{Weighted Jacobi solution scheme}
\label{sec: weighted Jacobi}

The global system of equations for the perturbed system, in which virtual modules are included in the model (Sec.~\ref{sec: virtual modules}) and contact conditions are accounted for by the predictor-corrector scheme (Sec.~\ref{sec: contact conditions}), reads:
	\begin{eqnarray}
	\label{eqn:pert local set of eqs}
	\bar{\bfm{K}} \bar{\bfm{u}} =\bar{\bfm{F}}^{\text{ext}}.
	\end{eqnarray}

Eq.~(\ref{eqn:pert local set of eqs}) is solved iteratively using the weighted Jacobi scheme \cite{Holobut17}. A single iteration $i\rightarrow i+1$ for module $p$ reads:
	\begin{eqnarray}
    \label{eqn: Jacobi}
	\bar{\bfm{u}}{}_p^{i+1}=\beta\bar{\bfm{D}}_p{}^{\!\!-1}\left(
	\bar{\bfm{F}}_p^{\text{ext}}-\bar{\bfm{R}}_p\bar{\bfm{u}}{}_p^i-\sum_q\bar{\bfm{K}}_{pq}^{12}\bar{\bfm{u}}{}_q^i\right) +(1-\beta)\bar{\bfm{u}}{}_p^i \,, \qquad
	\end{eqnarray}
	where $\bar{\bfm{D}}_p=\text{diag}\left(\sum_q \bar{\bfm{K}}_{pq}^{11}\right)$ and $\bar{\bfm{R}}_p=\left(\sum_q \bar{\bfm{K}}_{pq}^{11}\right)-\bar{\bfm{D}}_p$
	are the diagonal and the remainder parts of the respective stiffness sub-matrices, while $\beta=2/3$. Initially, we set $\bar{\bfm{u}}{}^0=\bfm{0}$ (alternatively, the solution for the non-perturbed state, if available, could be used instead of $\bfm{0}$, which would reduce the necessary number of iterations). 
	Note that in the iteration $i+1$ only the values of $\bar{\bfm{u}}$ from the iteration $i$ are used, and only those from $p$ and its direct neighbors $q$. Thus only local communication is involved and the memory complexity is constant.

In the present implementation, the weighted Jacobi procedure is initiated by the centroid module which sends an \emph{Init} message down the spanning tree, broadcasting the number of iterations to be done (see Sec.~\ref{sec: stability-of-simplified-case} for the centroid module and the spanning tree). Having received the \emph{Init} message, each \BB $B_p$ sends its initial 
vector $\bar{\bfm{u}}_p^0=\bfm{0}$ to all its neighbors and initializes its iteration counter, $\textrm{iter}_p=0$. 
In a given iteration $\textrm{iter}_p=i$, $B_p$ can receive from any of its neighbors $B_q$ a message containing $\bar{\bfm{u}}_q^i$ (displacements of $B_q$ calculated in iteration $i$), which is then stored in $B_p$'s buffer. When $B_p$ has received $\bar{\bfm{u}}_q^i$ from all its neighbors, it computes $\bar{\bfm{u}}_p^{i+1}$ (see Eq.~\ref{eqn: Jacobi}), increments its counter, $\textrm{iter}_p \gets \textrm{iter}_p+1$, and sends $\bar{\bfm{u}}_p^{i+1}$ to all its neighbors. The process continues until the prescribed number of iterations is reached.

\emph{Remark.} The weighted Jacobi procedure behaves like the Alpha local synchronizer~\cite{Raynal2013}.

\subsection{Convergence properties and possible improvements}
\label{sec:convergence_properties}
    The Weighted Jacobi scheme converges if the spectral radius of the iteration matrix $\bfm{C}_{\beta}=\bfm{I}-\beta\bar{\bfm{D}}^{-1}\bar{\bfm{K}}$ is less than 1:
    \begin{eqnarray}
    \rho(\bfm{C}_{\beta})=\text{max}(|\lambda_1|,\ldots, |\lambda_N|)<1,
    \end{eqnarray}
    where $\lambda_i$ are the eigenvalues of $\bfm{C}_{\beta}$. Although in the cases analyzed here convergence is achieved, the number of iterations is very high, which is a well-known drawback of the method (the convergence rate tends to 1 when the system grows~\cite{Strang2006}). 
    
    For the one-dimensional spring-in-series system of size $n$, we have analytically assessed the number of iterations necessary to attain an arbitrary relative error to be $O(n^2)$. This is also confirmed numerically in Sec.~\ref{sec: simulations and experiments} for more complex structures. The assessment shows the low efficiency of the scheme and underlies the complexities provided in Table~\ref{Tab: Complexities}.
    
\begin{table}
    \centering
\begin{tabular}{p{11em}|c|c|c}
 & $\mathbb{C}_{\text{C}}$ & $\mathbb{C}_{\text{T}}$, $\mathbb{C}_{\text{M}}$, $\mathbb{C}_{\text{SST}}$, $\mathbb{C}_{\text{MST}}$ & $\mathbb{C}_{\text{WJ}}$ \\
 \hline &&& \\[-0.9em]
   Execution time  & $O(d)$ & $O(\tilde{d})$ & $O(n^2)$\\[0.3em]
   No. of CPU operat./module  & $O(1)$ & $O(1)$ & $O(n^2)$\\[0.3em]
   Total no. of messages sent  & $O(n)$ & $O(n)$ & $O(n^3)$\\[0.3em]
   Memory usage per module  & $O(1)$ & $O(1)$& $O(1)$\vspace{1em}
\end{tabular}
\caption{Complexity assessments for subroutines of the algorithm. $\mathbb{C}_{\text{C}}$ refers to centroid selection, $\mathbb{C}_{\text{T}}$---to tree construction, $\mathbb{C}_{\text{M}}$---to finding the center of mass, $\mathbb{C}_{\text{SST}}$---to the simplified stability check, $\mathbb{C}_{\text{MST}}$---to the model-based stability check, and $\mathbb{C}_{\text{WJ}}$---to the weighted Jacobi procedure.}
\label{Tab: Complexities}
\end{table}

    The framework presented in this work is not restricted to the weighted Jacobi scheme though. Its efficiency can potentially be significantly improved by adapting another method to solving the considered contact problem in a distributed way. 
    We will briefly outline the three most promising directions.

\emph{Direction~1:} The Krylov subspace methods~\cite{Saad03} guarantee that the maximal number of iterations is at most equal to the number of degrees of freedom of the system, if the problem to be solved is linear. This can be further improved by appropriate preconditioning (see our preliminary study~\cite{LengiewiczIROS20}).
However, the need for global data aggregation and the non-linearities introduced by the contact problem require special treatment, which can deteriorate the time and memory efficiency.

\emph{Direction~2:} Multigrid techniques~\cite{BriggsHensonMcCormick2000} can more easily capture long-wave modes of the solution, which should improve the convergence rate. A special version must be devised, however, taking into account the contact conditions (like the one recently proposed~\cite{Wiesner2018}) and the specific computing architecture of the modular robot.

\emph{Direction~3:} The number of degrees of freedom of the system can be reduced by applying multi-scale methods or model order reduction techniques \cite{KerfridenEtAl2011, BeexEtAl2014}. However, it may be hard to find a suitable reduced space online efficiently. 

\section{Mechanical stability and overload check
}

\begin{figure}
\centering
\includegraphics[width=0.48\textwidth]{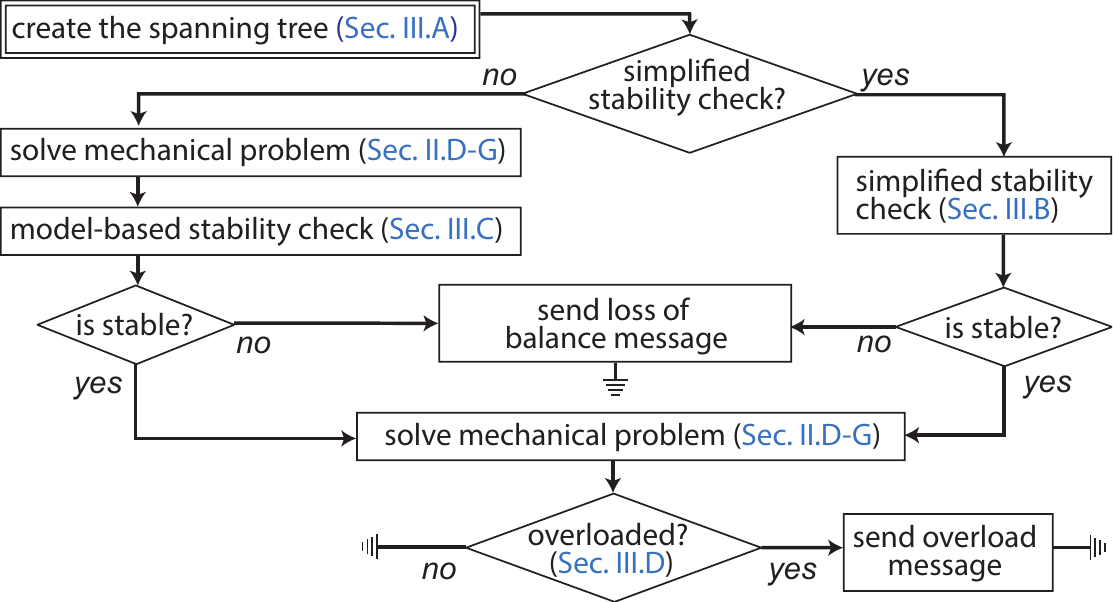}
	\caption{Flowchart of the algorithm. 
	}
	\label{fig:block_schematics}%
\end{figure}

Below we describe computational methods using which a robot can autonomously predict the two types of failure shown in Fig.~\ref{fig:failures}, one reconfiguration step ahead. The methods utilize a spanning tree, which is discussed first in Sec.~\ref{sec: spanning tree}. Sec.~\ref{sec: stability-of-simplified-case} describes a simple method of checking stability---without iterations, but restricted to robots standing on flat ground. Sec.~\ref{sec: stability for general case} discusses stability verification in the general case, using the iterative scheme of Sec.~\ref{sec: distributed prediction}. Finally, in Sec.~\ref{sec: overloading condition}, conditions for inter-modular connection breakage are presented, utilizing the results of the iterative scheme. The flowchart of the procedure is shown in Fig.~\ref{fig:block_schematics}.

\emph{Assessments of complexities} of the subroutines of the algorithm are presented in Table~\ref{Tab: Complexities}, where $n$ is the number of modules, $d$ is the radius of the connection graph of the robot, and $\tilde{d}$ is the depth of the spanning tree (usually, $\tilde{d}$ and $d$ are of the same order; see Sec.~\ref{sec: spanning tree} for more details).

\begin{figure}
\centering
\subfloat{\includegraphics[width=0.48\textwidth]{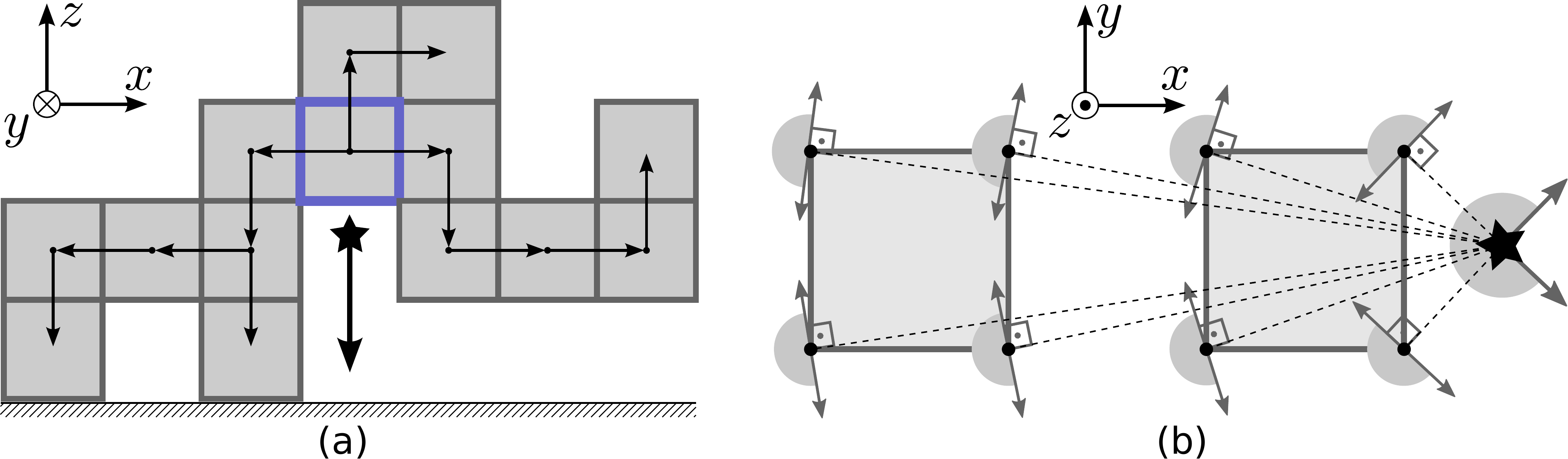}} 
\caption{(a) A modular robot (unstable) with a selected centroid (thick blue line), the center of mass (star) and a spanning tree (arrows). (b) Detection of whether a given point (star) is inside the convex hull of the support points. The condition is fulfilled if and only if all the straight angles (shaded) sum up to the full angle (they do not in this example).}
\label{fig:modular-robot-with-tree-and-convex-hull}
\end{figure}

\subsection{Spanning tree}
\label{sec: spanning tree}

A spanning tree allows efficient communication inside the robot. Its construction begins with a choice of the \emph{centroid} module, serving as the root, which is selected near the robot's topological center; see e.g. Fig.~\ref{fig:modular-robot-with-tree-and-convex-hull}a. This can be done automatically \cite{NazPiranda2016a}, but we chose the centroid manually in all examples.

The tree is extended to all modules, starting at the cen\-tro\-id which sends a \emph{Tree} message to its neighbours. When a module receives the \emph{Tree} message for the first time, it becomes a next-level node and sends the \emph{Tree} message further. This usually leads to the construction of BFS-like trees of quasi-optimal depth, without the need for synchronization. However, in rare cases the resulting tree may be far from optimal, with a Hamiltonian path over the structure as the worst-case scenario. The algorithms relying on the tree are then adversely affected.

\subsection{Loss of balance in a simplified case}
\label{sec: stability-of-simplified-case}

It is assumed that the robot is rigid and stands on flat ground under vertical gravity (Fig.~\ref{fig:modular-robot-with-tree-and-convex-hull}a).
The modules know their own masses and positions in a common Cartesian coordinate system, with $z=0$ being the ground level. They also simulate the behavior of their virtual neighbors (Sec.~\ref{sec: virtual modules}).

The stability check reduces to verifying that the center of mass of the robot lies over the convex hull of the points of support. If it does, the robot is stable, otherwise, it is not. The algorithm proceeds as follows.

\begin{enumerate}[leftmargin=*]
    \item[(a)] The center of mass of the robot is computed. Starting at the leaves of the spanning tree, each node sums up the masses of all its subbranches and its own, $m_i$, and likewise the weighted centers of mass of all its subbranches and its own, $m_i \bfm{X}_i$. The two sums are propagated to the parent node and the process continues. At the centroid, the center of mass of the robot is retrieved as $[X,Y,Z]=(\Sigma m_i \bfm{X}_i)/(\Sigma m_i)$.
    \item[(b)] $X$ and $Y$ are broadcast over the tree.
    \item[(c)] For any supporting module $i$, its \emph{safe angle range} $[\alpha_i, \beta_i]$ is determined as the sum of safe angle ranges of its corners. The $180^{\circ}$ safe angle range of a corner $\bfm{p}_j=[X_j,Y_j,0]$ is swept by the planar vector $[X_j-X,Y_j-Y]$ when turning $90^{\circ}$ left or right. It covers those directions in which the structure cannot tilt; see Fig.~\ref{fig:modular-robot-with-tree-and-convex-hull}b. The safe angle range of any corner $\bfm{p}_j=[X_j,Y_j,Z_j\neq 0]$ is assumed to be empty.
    \item[(d)] The safe angle ranges are summed up over the tree, just like masses were in step (a). The summation always gives a single interval, because all considered ranges are either empty or not less than the straight angle.
    \item[(e)] The structure is stable if and only if the aggregated angle range at the centroid equals the full angle.
\end{enumerate}

\subsection{Loss of balance in the model-based approach}
\label{sec: stability for general case}

In the general case with arbitrarily placed supports, there is no simple method to predict stability. Sometimes, if a model of the robot is simple, like under the rigid-body or elasto-static assumptions used here, there may even be no unique answer. Since more accurate modeling goes beyond the scope of the present paper, we will show how to utilize the proposed elasto-static model with contact and the iterative solution scheme to check stability in more general cases. The method, however, does not assure finding the correct solution in difficult cases (e.g., when the solution is non-unique by definition).

The method is based on the observation that, when the iterative solution scheme has converged, the local state of the contact conditions is fully determined. It is only necessary to check whether active supports prevent rigid-body rotation of the structure (at least three noncollinear support points must be active), which is achieved by the following procedure:

\begin{enumerate}[leftmargin=*]
    \item[(a)] Solve the mechanical problem (see Sec.~\ref{sec: distributed prediction}) and initiate the stability check (the spanning tree is used again).
    \item[(b)] For every module,
determine the set of its active corner points by checking the contact conditions (Sec.~\ref{sec: contact conditions}):
    \begin{itemize}
        \item No contact $\to$ return the empty set. 
        \item Tilting in two directions $\to$ return a single corner---the common point of the two edges of rotation.
        \item Tilting in one direction $\to$ return two corners---the end points of the edge of rotation.
        \item Contact \& no tilting $\to$ return a special `stable' state.
    \end{itemize}
    \item[(c)] Aggregate information starting at the leaves and moving up the spanning tree towards the centroid:
    \begin{itemize}
        \item At each module, sum the sets of active points of its subbranches and its own, obtaining set $S$. By convention, adding any set to `stable' gives `stable'.
        \item If the points in $S$ are noncollinear then set $S=\text{`stable'}$.
        \item If multiple points in $S$ are collinear then leave only two.
        \item Pass $S$ up the spanning tree.
    \end{itemize}
    \item[(d)] The stability check ends at the centroid,
    with the result being either `stable' or not.
\end{enumerate}

Excluding the expensive phase of determining active supports (weighted Jacobi iterations), the complexity of the approach is the same as that of the simplified case; see Table~\ref{Tab: Complexities}. The memory complexity per module is constant because each returned set of points has at most two elements.

\subsection{Overloading of inter-modular connections}
\label{sec: overloading condition}

Connection overloading is checked when the iterative scheme has sufficiently converged and after checking that the structure is stable. The forces and torques which act between module $p$ and its neighbor $q$ are predicted as follows:
\begin{eqnarray}
\nonumber{}&&[f_x, f_y, f_z, m_x, m_y, m_z]^{\text{T}} = \frac{1}{2}\bfm{\hat{R}}_{pq}^{\text{T}}(\bfm{F}_{pq}-\bfm{F}_{qp}) =\\
&&\qquad=\frac{1}{2}\bfm{\hat{R}}_{pq}^{\text{T}}(\bar{\bfm{K}}^{11}_{pq}\bar{\bfm{u}}_p + \bar{\bfm{K}}^{12}_{pq}\bar{\bfm{u}}_q - \bar{\bfm{K}}^{11}_{qp}\bar{\bfm{u}}_q - \bar{\bfm{K}}^{12}_{qp}\bar{\bfm{u}}_p)\,,\qquad \label{eq: mid-stresses}
\end{eqnarray}
where $\bfm{\hat{R}}_{pq}^{\text{T}}$ rotates the resulting vector into a coordinate system in which axial forces are aligned with the $z$ axis.

To avoid connection breakage, the tensile force $f_z$ and torques $m_x$ and $m_y$, computed in Eq.~(\ref{eq: mid-stresses}) in the middle of the connection, must not overpower the magnetic forces $F^{\textrm{max}}$ binding the modules. (Shear and torsion are omitted, because \emph{Blinky Blocks}' connectors are assumed to be resistant to those modes of breakage.) The vertical and lateral connections of \BBs differ, so that $F^{\textrm{max}}$ can take two values; see Table~\ref{tab: BlinkyBlock material parameters}. 
The safety condition for both tension and bending is:
\begin{equation} \label{eq: safety conditions}
F^{\textrm{max}} > 2\cdot\max(|m_x|,|m_y|)/L + f_z \,.
\end{equation}
The check is performed for all connections and the results are aggregated by the centroid over the spanning tree.

\section{Implementation, simulations and experiments}

\subsection{Implementation details}
\label{sec:Implementation details}

The procedures have been implemented and tested in the integrated environment developed at FEMTO-ST\footnote{Programmable Matter project at FEMTO-ST: \url{https://projects.femto-st.fr/programmable-matter/}.}. It consists of the virtual test bed \VS~\cite{dagstuhlVS} and the reconfigurable modular robot \BBs \cite{GoldsteinCHI2011}, so the same implementation could be executed on both platforms. The software was adjusted to be compatible with the real \BBs Version~1 hardware: reduced floating-point precision of \SI{4}{Bytes} was used and the maximum message size was set to \SI{17}{Bytes} (messages containing $6\times 4$ Byte-long vectors were split in half).

The program's flowchart is shown in Fig.~\ref{fig:block_schematics}, and the consecutive steps of the algorithm are discussed in the previous sections. The choice between the simplified (Sec.~\ref{sec: stability-of-simplified-case}) and the full (Sec.~\ref{sec: stability for general case}) stability check to be performed is preset by the user.
In the case of the \BBs hardware, a preliminary step is additionally performed, in which the same main program is loaded into each \BB, 
and a common coordinate system for a given configuration of \BBs is propagated, starting from a special block with preset coordinates.

\begin{figure}
\centering
\subfloat{ \includegraphics[width=0.47\textwidth]{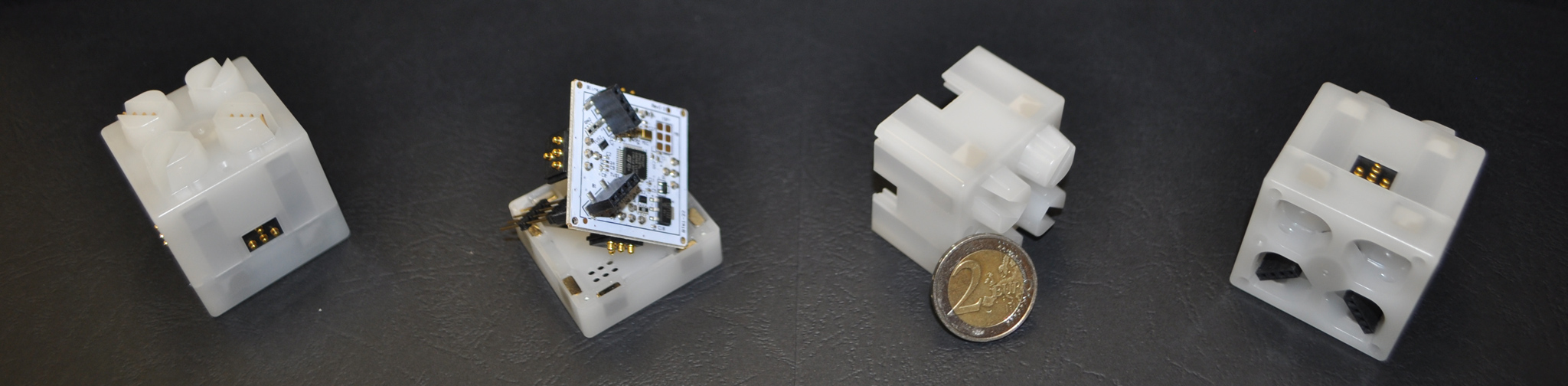} }
	\caption{\BBs: functional ones (left and right), and two pieces of a dismantled one with a top view of the motherboard (in the middle).}
	\label{fig:fig1}%
\end{figure}

The material parameters of the \BB model described in Sec.~\ref{sec: Standard 3D frame model} are provided in Table~\ref{tab: BlinkyBlock material parameters}. All have been assessed experimentally, except Young's modulus which was chosen arbitrarily (its exact value is not essential for assessing overload and stability). The dimensions and mass have been measured directly. Connection strengths have been obtained in a simplified manner, by finding the maximum number of \BBs that their magnets could hold hanging in a vertical alignment. The top/bottom and lateral connection strengths differ because the former is produced by a Lego-like system reinforced with a central magnet, and the latter by 4 magnets placed in the corners of each face; see also Fig.~\ref{fig:fig1}. 

\subsection{Simulations and experiments}
\label{sec: simulations and experiments}

\begin{figure*}[th!]
\centering
\subfloat{ \includegraphics[width=0.473\textwidth]{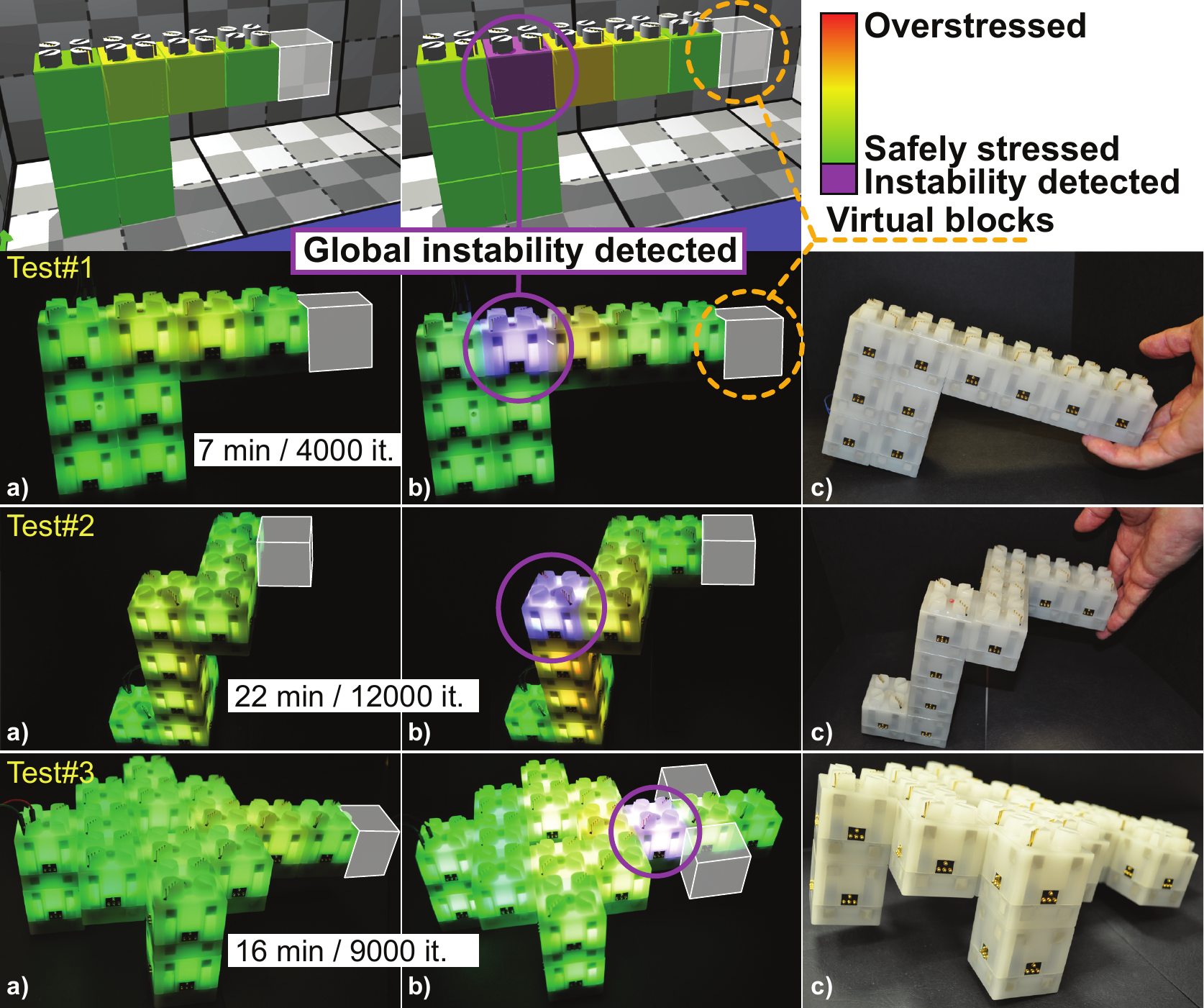} } \hspace{0.02\textwidth}
\subfloat{ \includegraphics[width=0.473\textwidth]{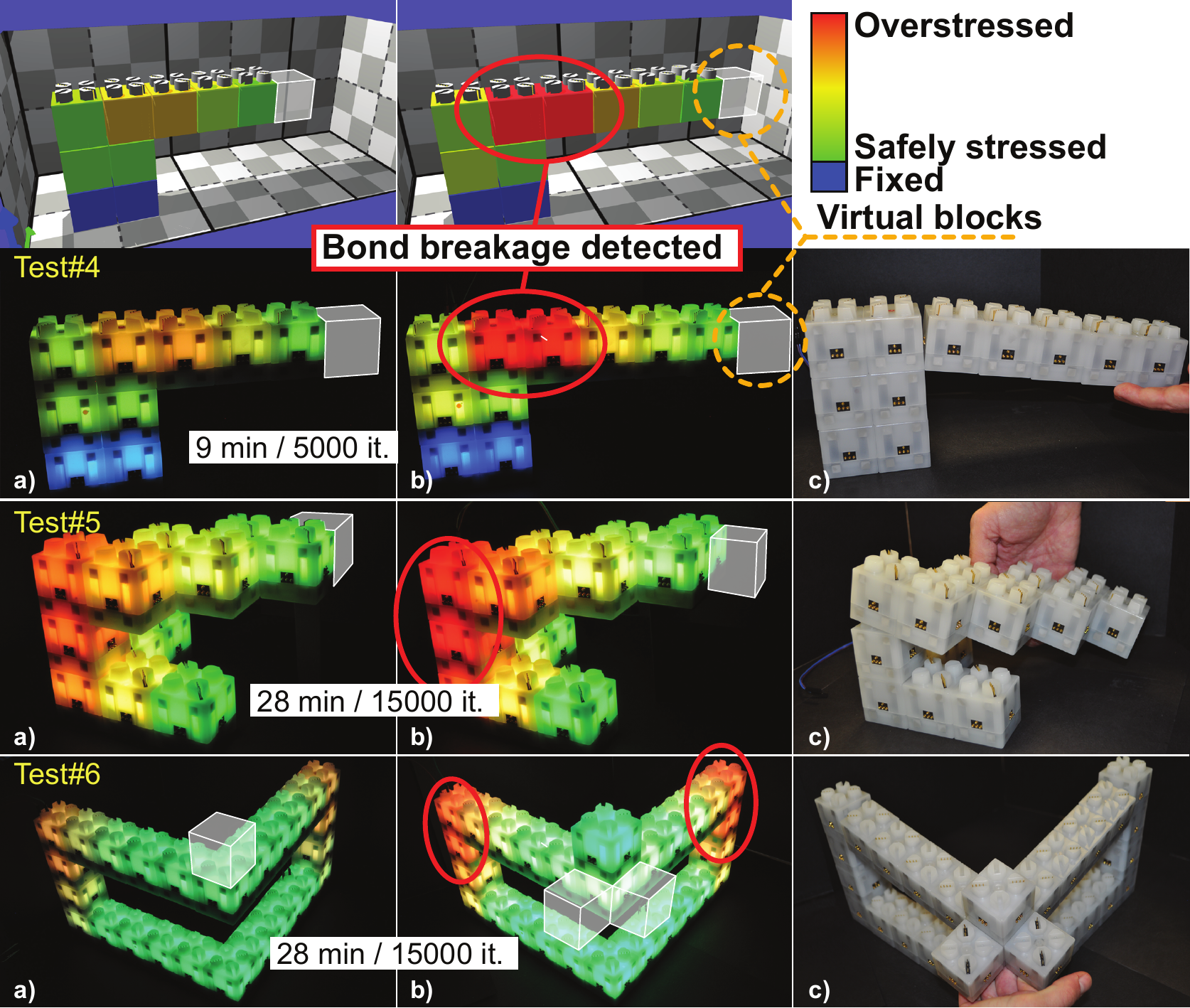} }
	\caption{Experiments in \VS and on \BBs. Computation times and iteration numbers are shown in insets.} 
	\label{fig:experiments}%
\end{figure*}

Six different modular configurations and failure scenarios were investigated (see Fig.~\ref{fig:experiments}), both in \VS and on \BBs\footnote[2]{Videos of selected experiments: \href{https://youtu.be/d3aE8GjbYd8}{https://youtu.be/d3aE8GjbYd8}}. The sizes of structures ranged from 8 modules in test~\#1(a) to 29 modules in test~\#6(c). Experiments on substantially larger structures would be difficult due to the high computational complexity of the algorithm combined with the limited processing and communication speed of the Version 1 of \BBs. In the physical experiments, reconfiguration of \BBs was done by manually attaching new modules to an existing structure. 
In all the presented cases, the model-based analysis was involved (addressing both overloading and loss of stability), which required execution of the weighted Jacobi iterative scheme. 
Additionally, in the loss-of-stability scenarios (Fig.~\ref{fig:experiments} \#1-\#3), the results obtained with the simplified and the model-based stability checks were successfully cross-validated.

Because it is in general difficult to automatically assess the necessary number of weighted Jacobi iterations, this number was adjusted manually case by case. The criterion was to make the number of iterations possibly low while obtaining correct predictions at the same time. The problem of how the necessary number of iterations scales with the system size in a stable case is briefly discussed later. In an unstable case, this number is expected to be much higher. Following the conclusion in the final Remark of Sec.~\ref{sec: contact conditions}, in unstable cases we stopped computations just after all contact states stabilized, but before numerical convergence was achieved.

Each of the six tests in Fig.~\ref{fig:experiments} shows the results of execution of the program for two consecutive construction steps of a particular \BBs structure. In every figure, the first construction step~(a$\to$b) is designed to be mechanically safe, while the second one~(b$\to$c) to result in failure, which is then demonstrated in the third part of the figure~(c). Additionally, in the top row, \VS results are shown for the tests \#1 and \#4. The tests are split into loss-of-stability (left column) and overloading (right column) scenarios. From top to bottom, the scenarios are ordered by complexity, i.e., 2D cases in tests~\#1~\&~\#4, 3D cases in tests~\#2~\&~\#5, and 3D cases with more complex connection topologies in tests~\#3~\&~\#6.

The results of calculations are displayed using colors: the color of a block corresponds to the highest tensile/bending stress level in any of its connections, as given by the right hand side of Eq.~(\ref{eq: safety conditions}). Green to orange colors represent the safe stress range, while red indicates potentially overstressed connections. \BBs were programmed to blink in red when tensile stresses in some of their connections exceeded the critical level, while global imbalance of a structure was signaled by the centroid module blinking in purple. Blue \BBs are fixed---they are attached to the floor.

In all tests except \#6, the predictions are confirmed by physical experiments. In test~\#6, breakage is correctly predicted but ill-localized. This can be observed in Fig.~\ref{fig:experiments}-\#6(b) which indicates breakage of the pillars, while the actual breakage occurs as it is shown in Fig.~\ref{fig:experiments}-\#6(c).  There are two possible reasons for the observed discrepancy.
The first one is that the assumed mechanical model of the modular robot is too simple. The second one is the omission of twisting torques from the adopted criterion of breakage. It was also very difficult to keep the structures \#6(a) and \#6(b) operational---an effect which was not expected. In both cases, weight-induced deformations caused separation of electrical connectors, despite the structures did not break. This necessitated using additional supports just to perform computations. We view test~\#6 as one of benchmark cases for future research on more accurate models and failure criteria.

\begin{figure}
\centering
\subfloat{\small(a)\hspace{-1.4em}\includegraphics[height=0.164\textwidth]{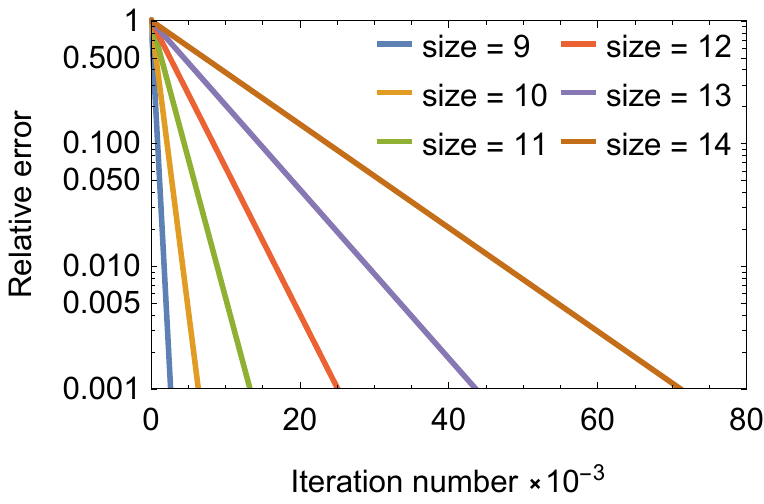}\label{fig:armfixed convergence}}%
\subfloat{\small(b)\hspace{-1.4em}\includegraphics[height=0.162\textwidth]{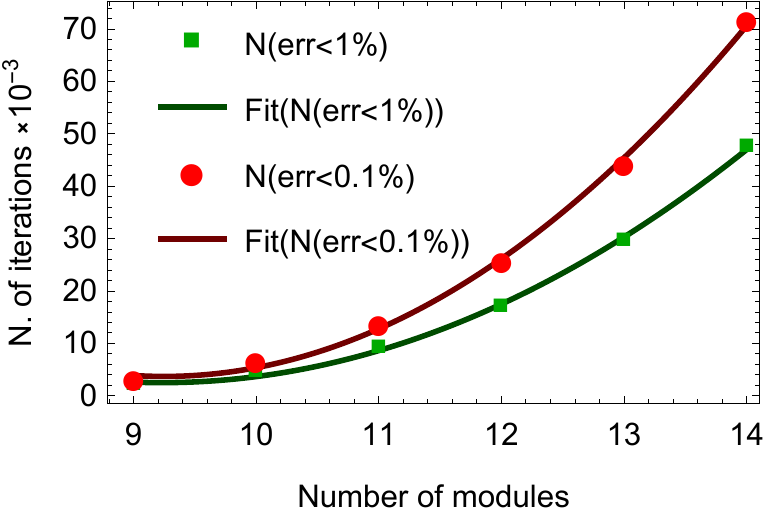}\label{fig:armfixed nof iterations}}
	\caption{Number of weighted Jacobi iterations necessary to attain a given accuracy for fixed-arms of different sizes. Size 10 refers to the example in Fig.~\ref{fig:experiments}-\#4(a); other sizes have shorter/longer arms.
(a) Convergence characteristics. (b) A quadratic polynomial fit for two accuracy levels.}
	\label{fig:armfixed conv}%
\end{figure}

\emph{CPU time and convergence properties.}
Computing $\bar{\bfm{u}}^i_p$ and exchanging messages with neighbours takes a \BB a nearly constant time $T \approx $\SI{110.5}{\milli \second}  ($9.05$ runs per second). Because communication is local, $T$ is also the global time of a single weighted Jacobi iteration, independent of the configuration. Since the time cost of the other steps of the algorithm is negligible (see Tab.~\ref{Tab: Complexities}), the overall execution time can be assessed by multiplying the number of iterations by~$T$.

The number of iterations needed to attain a given accuracy greatly depends on the system's configuration and generally grows with the number of modules $n$. Assessment of the number of iterations is generally not straightforward, even without considering unilateral contact conditions; see also Sec.~\ref{sec:convergence_properties}. In Fig.~\ref{fig:armfixed conv} we demonstrate the expected trends for a given family of configurations. Fig.~\ref{fig:armfixed convergence} shows linear convergence of the relative error $\|\bar{\bfm{u}}^i-\bar{\bfm{u}}^{*}\|/\|\bar{\bfm{u}}^{*}\|$ as the number of iterations $i$ grows, where $\bar{\bfm{u}}^{*}$ is the numerically exact solution. Fig.~\ref{fig:armfixed nof iterations} presents the necessary numbers of iterations from Fig.~\ref{fig:armfixed convergence} for two example relative errors, displaying quadratic growth with $n$ and confirming the assessments in Tab.~\ref{Tab: Complexities}.

\section{Conclusions and future research}

We presented a distributed algorithm for checking if a modular robot will retain its mechanical integrity and stability after new modules are attached to it at prescribed positions. The algorithm can be used to assess the mechanical safety of a reconfiguration step planned by a self-reconfigurable robot. The procedure is designed to run on the modular robot itself, and we have verified its predictions through tests in the dedicated simulator \VS and on the real modular system \BBs. To our knowledge, this is the first time three-dimensional modular-robotic structures compute their mechanical state in a fully distributed manner.

The algorithm can be improved towards: adopting faster iterative schemes, as discussed in Sec.~\ref{sec:convergence_properties} and tried in \cite{LengiewiczIROS20}; extending the application range to soft modular robots; checking the construction/reconfiguration several steps ahead; as well as addressing other module geometries and broader module-support contact scenarios. Future experimental validation will use the currently produced new version of Blinky Blocks with faster CPUs and communication, and possibly quasi-spherical catoms~\cite{PB2018} having up to 12 neighbors per module and electrostatic connectors.


\vspace{1em}\noindent\emph{Acknowledgement.} This work was partially supported by the EU Horizon 2020 Marie Sklodowska Curie Individual Fellowship \emph{MOrPhEM} (grant No.\ 800150), by the NCN Project ``Micromechanics of Programmable Matter'' (grant No. 2011/03/D/ST8/04089),  by the ANR (ANR-16-CE33-0022-02), the French Investissements d'Avenir program, ISITE-BFC project (ANR-15-IDEX-03), EIPHI Graduate School (contract ANR-17-EURE-0002), Mobilitech project and EU Horizon 2020 research and innovation programme (grant No. 811099 TWINNING Project DRIVEN for the Univ. of Luxembourg).
%
%
\bibliography{PM_bibliography}
\bibliographystyle{ieeetr}

\end{document}